\documentclass[11pt]{article}

\usepackage[final]{acl}

\usepackage{times}
\usepackage{latexsym}

\usepackage{inconsolata}

\usepackage[utf8]{inputenc} 
\usepackage[T1]{fontenc}    
\usepackage{hyperref}       
\usepackage{url}            
\usepackage{booktabs}       
\usepackage{amsfonts}       
\usepackage{nicefrac}       
\usepackage{microtype}      
\usepackage[table]{xcolor}         
\usepackage{tabularx}
\usepackage{multirow}
\usepackage{graphicx}
\usepackage{amsmath}
\usepackage{array}
\usepackage{pifont}
\usepackage{makecell}
\usepackage{wrapfig,lipsum,booktabs}
\usepackage{colortbl}
\usepackage[most]{tcolorbox}

\newcommand{\tablestyle}[2]{\setlength{\tabcolsep}{#1}\renewcommand{\arraystretch}{#2}\centering\footnotesize}
\newlength\savewidth\newcommand\shline{\noalign{\global\savewidth\arrayrulewidth
  \global\arrayrulewidth .8pt}\hline\noalign{\global\arrayrulewidth\savewidth}}  

\definecolor{maroon}{cmyk}{0,0.1,0.01,0.01}
\definecolor{blue}{cmyk}{0.95,0.0,0.2,0.2}
\definecolor{yellow}{cmyk}{0.01,0.0,0.2,0.01}
\definecolor{lightblue}{cmyk}{0.1,0.0,0.02,0.02}
\definecolor{case_verb}{HTML}{fbde84}
\definecolor{case_adj}{HTML}{cccdff}
\definecolor{case_noun}{HTML}{bfeaf1}
\definecolor{case_ff}{HTML}{e65352}
\definecolor{case_error}{HTML}{ffff00}
\definecolor{darkgreen}{RGB}{51,181,41}
\definecolor{darkorange}{RGB}{252,135,62}
\definecolor{t_green}{HTML}{f1f2e4}
\definecolor{LIGHT_BLUE}{HTML}{cce4fe}
\definecolor{LIGHT_RED}{HTML}{f1b9b8}
\definecolor{LIGHT_YELLOW}{HTML}{f1f58a}
\definecolor{LIGHT_GREEN}{HTML}{f1f2e4}
\definecolor{LIGHT_PURPLE}{HTML}{b6a7b9}
\definecolor{lightgray}{gray}{0.95}
\usepackage{enumitem}
\setlist[itemize]{leftmargin=*}
\setlist[enumerate]{leftmargin=*}
\lstdefinestyle{prompt}{
    basicstyle=\ttfamily\fontsize{7pt}{8pt}\selectfont,
    frame=none,
    breaklines=true,
    backgroundcolor=\color{lightgray},
    breakatwhitespace=true,
    breakindent=0pt,
    escapeinside={(*@}{@*)},
    numbers=none,
    numbersep=5pt,
    xleftmargin=5pt,
}
\tcbset{
  aibox/.style={
    top=10pt,
    colback=white,
    colframe=black,
    colbacktitle=black,
    enhanced,
    center,
    attach boxed title to top left={yshift=-0.1in,xshift=0.15in},
    boxed title style={boxrule=0pt,colframe=white,},
  }
}
\newtcolorbox{AIbox}[2][]{aibox, title=#2,#1}

\title{OPT-BENCH: Evaluating the Iterative Self-Optimization of LLM Agents in Large-Scale Search Spaces}

\author{ Xiaozhe Li$^{1}$, Jixuan Chen$^{2,}$$^{5}$, Xinyu Fang$^{2,}$$^{4}$, Shengyuan Ding$^{2,}$$^{3}$,\\ \bf Haodong Duan$^{2}$, Qingwen Liu$^{1}$$^{\dag}$, Kai Chen$^{2}$\\
Tongji University$^{1}$, Shanghai AI Lab$^{2}$, Fudan University$^{3}$, Zhejiang University$^{4}$,\\ University of California San Diego$^{5}$\\
}

\begin{document}
\twocolumn[{%
\renewcommand\twocolumn[1][]{#1}%
\maketitle
\vspace{-20mm}

\begin{center}
    \centering
    \captionsetup{type=figure}
    \vspace{10mm}
    \includegraphics[width=1\linewidth]{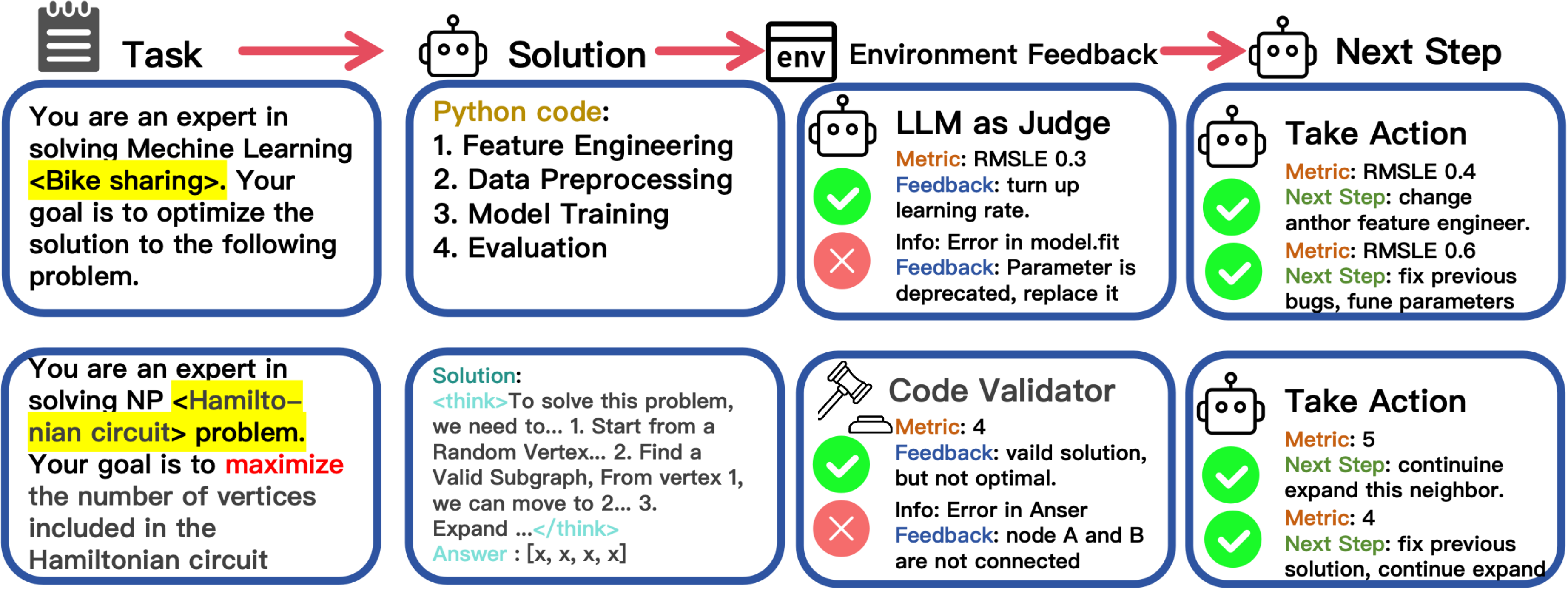}
    \caption{
    \textbf{OPT-BENCH framework.} 
    This framework evaluates \textbf{Iterative Self-Optimization} by integrating two distinct reasoning modalities: \textbf{Continuous Parametric Optimization} for machine learning tasks (Top), and \textbf{Discrete Combinatorial Reasoning} for NP-hard problems (Bottom). In both settings, the agent leverages \textbf{Environmental Feedback} to guide its trajectory of improvement and debugging, progressively bridging the gap between initial hypotheses and optimal solutions through intrinsic reasoning.}
    \label{fig:spotligh}
    \vspace{3mm}
\end{center}%
}
]

\footnotetext[1]{$^{\dag}$ represents the corresponding author, email: qliu@tongji.edu.cn.}
\begin{abstract}
Large Language Models (LLMs) have demonstrated remarkable capabilities in reasoning and tool use. However, the fundamental cognitive faculties essential for problem-solving—perception, reasoning, and memory—remain the stable core of intelligence. Unlike memorizing specific patterns, humans succeed in novel environments by applying these intrinsic faculties to adapt and optimize. Yet, whether LLMs possess this essential capacity—namely, the ability to continuously refine solutions in response to dynamic environmental feedback—remains underexplored.
To address this challenge, we introduce \textbf{OPT-BENCH}, a benchmark for evaluating self-improvement capabilities in large-scale search spaces. By combining 20 machine learning tasks with 10 classic NP-hard problems, OPT-BENCH provides a rigorous setting to assess whether agents can adapt through intrinsic self-reflection rather than rote tool application.
We further propose \textbf{OPT-Agent}, a framework that emulates human-like cognitive adaptation. It operates via a general perception--memory--reasoning loop, iteratively refining solutions based on environmental feedback.
Through extensive experiments on 19 LLMs from 7 model families, including reasoning models, general models, and open-source models ranging from 3B to 235B parameters, we demonstrate stronger models are more effective at leveraging feedback signals for self-improvement. However, this upper-bound adaptability remains fundamentally constrained by the models' base capacity, and even the most advanced LLMs still fall short of human expert performance.
\end{abstract}

\section{Introduction}

The advent of Large Language Models (LLMs)~\cite{o1-system-card, grattafiori2024llama, guo2025deepseek} has revolutionized artificial intelligence, demonstrating exceptional performance across a wide range of tasks~\cite{brown2020language, ouyang2022training, achiam2023gpt, chowdhery2023palm, touvron2023llama, team2024gemini}. While specific interaction tools evolve rapidly, the fundamental cognitive faculties required for problem-solving—\textit{perception}, \textit{reasoning}, and \textit{memory}—remain the stable core of intelligence. Humans succeed in novel environments not by memorizing specific interfaces, but by applying these intrinsic principles to adapt and optimize. However, determining whether LLMs possess this essential capacity for \textbf{iterative self-optimization}—continuously refining solutions in response to \textbf{dynamic environmental feedback}—remains an underexplored frontier.

Current benchmarks primarily measure a model's ability to generate correct responses in a single pass~\cite{fan2023nphardeval, lin2024criticbench}, overlooking the essential human capability of learning from experience. Humans routinely refine reasoning by integrating feedback—scientists adjust hypotheses, students revise strategies, and chess players improve tactics based on past outcomes. Yet, existing evaluations lack a unified framework to assess this \textit{trajectory} of self-improvement, leaving a critical gap in understanding how agents navigate complex search spaces through intrinsic reflection rather than rote memorization.

To address this limitation, we introduce \textbf{OPT-BENCH}, a comprehensive benchmark designed to probe the limits of LLM agents in large-scale search space self-optimization driven by environmental signals. OPT-BENCH juxtaposes two distinct feedback landscapes:
(1) \textbf{20 Real-world Machine Learning (ML) tasks} (sourced from Kaggle), representing \textit{continuous optimization spaces} where feedback provides noisy but directional gradients (e.g., improving validation accuracy); and 
(2) \textbf{10 Classical NP-hard problems}, representing \textit{discrete combinatorial spaces} characterized by brittle constraints and local optima.
This deliberate combination challenges agents to demonstrate versatile cognitive adaptability: employing \textbf{inductive reasoning} to tune hyperparameters in ML tasks, while switching to \textbf{deductive logic} to construct valid structures in NP problems. To evaluate performance relative to human cognition, we curate gold-standard solutions from Kaggle leaderboards for ML tasks and implement \textbf{Human-expert Heuristics} for NP tasks, establishing a baseline rooted in expert reasoning patterns rather than brute-force computation.

To facilitate rigorous evaluation, we present \textbf{OPT-Agent}, a framework designed to emulate \textbf{human-like cognitive adaptation}. Unlike complex agent architectures that rely heavily on engineered prompts or external solvers, OPT-Agent employs a \textbf{perception--memory--reasoning loop}. It generates solutions, validates them against the environment, and iteratively refines them by retrieving and analyzing \textit{accumulated environmental signals}.

Using this benchmark, we conduct extensive experiments on 19 LLMs from 7 model families. Our analysis uncovers a "Scaling Law of Self-Improvement": stronger base models are significantly more effective at leveraging historical feedback to optimize solutions. However, we reveal a critical \textbf{cognitive divergence}: while historical context effectively drives optimization in continuous ML domains, its benefit is often constrained in discrete NP tasks. Even frontier models (e.g., GPT-4o) struggle with \textit{incremental refinement} in combinatorial spaces, frequently resetting solutions rather than repairing them—highlighting a persistent gap compared to human expert performance.

In summary, our contributions are:
\begin{itemize}[itemsep=4pt, topsep=0pt, parsep=0pt]
    \item We propose \textbf{OPT-BENCH}, a benchmark that juxtaposes continuous (ML) and discrete (NP) optimization tasks to evaluate the \textit{intrinsic self-optimization} capabilities of LLMs, moving beyond rote tool application to assess dynamic adaptability.
    \item We introduce \textbf{OPT-Agent}, an evaluation framework that emulates the human perception-memory-reasoning loop, enabling a rigorous assessment of how models leverage historical environmental feedback to refine solutions iteratively.
    \item We provide a comprehensive analysis of 19 LLMs, revealing that while strong models excel at inductive optimization in continuous spaces, they face fundamental reasoning barriers in discrete combinatorial refinement, marking a key direction for future AGI research.
\end{itemize}
\begin{figure*}[t!]
    \centering
    \includegraphics[width=\linewidth]{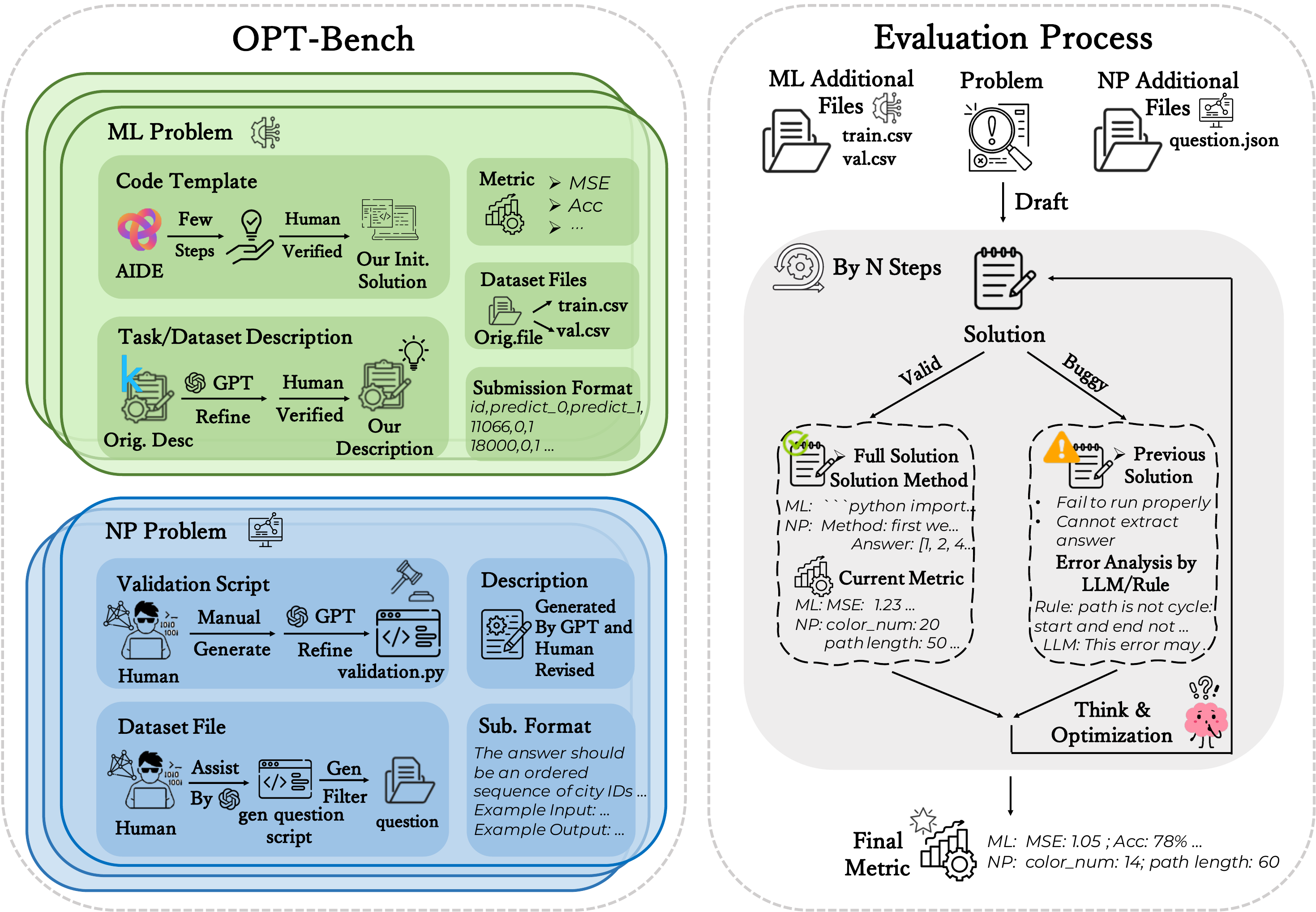}
    \caption{\textbf{Overview of the OPT-BENCH dataset and OPT-Agent Framework.} The left panel illustrates the data structure of OPT-Bench, encompassing ML and NP problems. Each module includes problem definitions, dataset files, validation script (NP), evaluation metrics, and submission formats, integrating human-verified initial solutions and LLM-assisted refinement. The right panel details the evaluation workflow, where solutions are iteratively generated and validated in steps. Valid solutions proceed to metric calculation, while buggy solutions trigger error analysis via LLM or rule-based diagnostics, followed by iterative optimization. This framework enables systematic and automated assessment of LLMs' optimization capabilities across diverse problem domains.}
    \label{fig:pipeline}
    \vspace{-1.5em}
\end{figure*}
\section{OPT-BENCH}
To rigorously evaluate adaptive intelligence across diverse feedback landscapes, OPT-BENCH integrates 30 distinct tasks: 20 machine learning (ML) challenges and 10 classical NP-hard problems. This composition is intentionally designed to juxtaposition two fundamental modes of reasoning:
(1) \textbf{Continuous Inductive Optimization} (ML tasks): utilizing noisy, directional feedback (e.g., accuracy metrics) to tune continuous hyperparameters in applications like sales forecasting and sentiment analysis.
(2) \textbf{Discrete Deductive Reasoning} (NP tasks): navigating brittle, combinatorial search spaces with strict constraints, covering graph theory and resource allocation problems such as Hamiltonian cycle and TSP.
These problems are selected for their computational intractability at scale, forcing agents to rely on heuristic planning rather than brute-force memorization. See Appendix A for the complete task list.

\subsection{Dataset Curation and Analysis}
\label{sec:dataset_curation}
As illustrated in Figure~\ref{fig:pipeline}, the preparation of ML tasks ensures a rigorous testbed for real-world engineering capabilities. We collect task descriptions from Kaggle, refine them via GPT-4o, and verify them through human experts to ensure ambiguity-free instructions. Evaluation metrics are strictly defined to provide objective feedback signals. To simulate a realistic optimization starting point, we generate an initial solution using the AIDE agent~\cite{aide}, followed by refinement from four PhD-level experts. This ensures the initial state is functional but suboptimal, providing ample headroom for iterative improvement. We utilize Kaggle leaderboard gold medal solutions as the \textbf{Human Expert Baseline}, representing the upper bound of domain-specific engineering.

For NP tasks, the benchmark focuses on 10 classical problems encapsulated in JSON format. Unlike ML tasks where the solution is a script, NP tasks require the construction of valid discrete structures (e.g., a specific node sequence). We provide detailed goal definitions and example I/O to guide the model's deductive process. Validity is enforced by a rule-based script \texttt{validation.py}, which acts as the "Environment," providing binary feedback (Valid/Invalid) and specific error messages (e.g., "Node visited twice"). Crucially, to establish a meaningful comparison for intrinsic reasoning, we implement \textbf{Human-Exper Heuristics} (e.g., simulated annealing algorithm) as the baseline. This serves as a proxy for expert human reasoning patterns, distinct from the theoretical optima found by industrial solvers.

\begin{figure*}[t]
    \centering
    \includegraphics[width=0.95\linewidth]{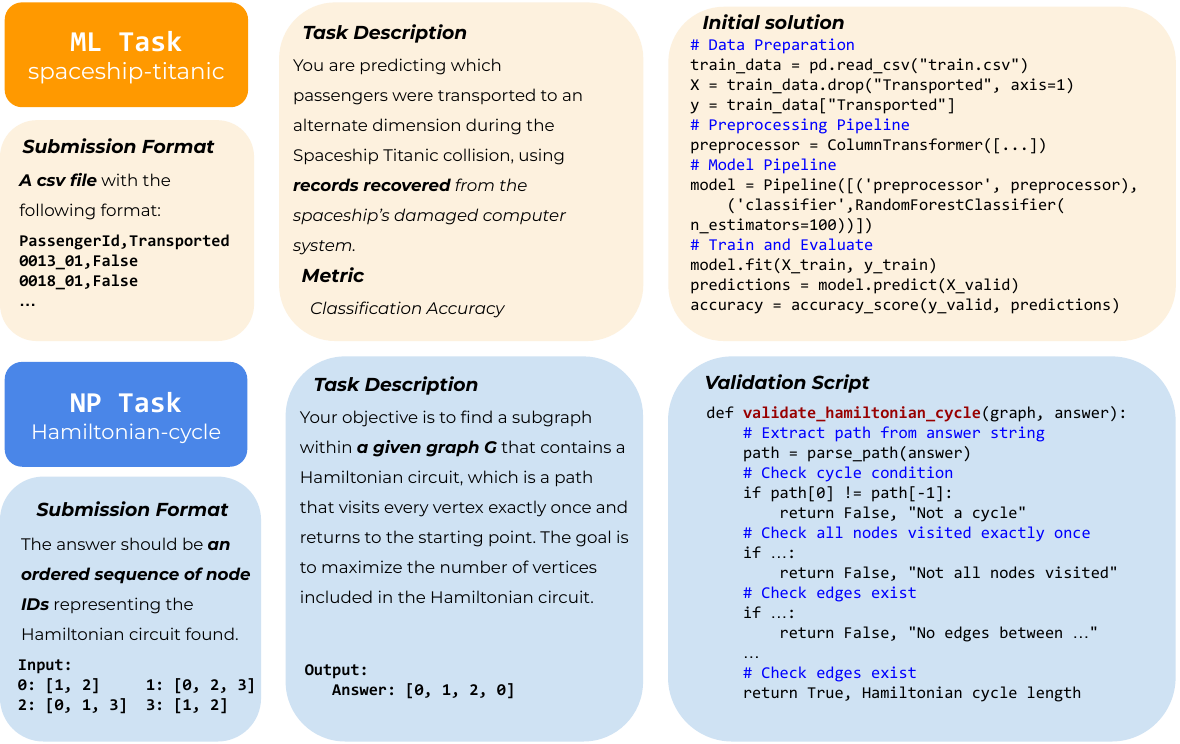}
    \caption{\textbf{Specific cases from OPT-BENCH}. Take the spaceship titanic classification task and the Hamiltonian cycle optimization problem as representative examples.}
    \label{fig:data_case}
    \vspace{-1.5em}
\end{figure*}

Each sample in OPT-BENCH comprises the following elements (see Figure~\ref{fig:data_case}): 
\begin{itemize}
    \item \textbf{Task Description}: Defines the optimization landscape, including background context and specific objectives (e.g., minimize path length).
    \item \textbf{Dataset Specifications}: ML tasks provide CSVs with feature descriptions; NP tasks provide JSON instances of varying topological complexity.
    \item \textbf{Submission Formats}: Stipulates the required output structure, ensuring consistent parsing for automated evaluation.
    \item \textbf{Initial Solution (Cold Start)}: A functional but suboptimal baseline (script for ML, valid path for NP) provided to the agent to kickstart the optimization trajectory.
    \item \textbf{Environmental Feedback Mechanism}: For ML, this involves execution on a hold-out test set; for NP, \texttt{validation.py} verifies constraints and computes metrics. This mechanism simulates dynamic environmental signals.
    \item \textbf{Human/Expert Baseline}: Kaggle Gold Solutions for ML, and Human-Proxy Heuristics for NP, providing a "reasoning ceiling" for comparison.
\end{itemize}

\subsection{OPT-Agent Workflow}
To better evaluate the intrinsic self-optimization capabilities of LLMs, we introduce \textbf{OPT-Agent}, a framework inspired by AlphaEvolve~\cite{2025alphaevolve} and cognitive Chain-of-Thought theories. Instead of relying on elaborate prompt engineering, OPT-Agent implements a fundamental \textbf{Perception--Memory--Reasoning Loop}. As illustrated in Figure~\ref{fig:pipeline}, it iteratively refines solutions by accumulating and analyzing environmental signals.

\begin{itemize}
    \item \textbf{Drafting (Initialization)}: The agent enters the environment by generating an initial hypothesis. For ML tasks, it synthesizes a Python training script; for NP tasks, it uses \textbf{deductive reasoning} to directly construct a solution structure (e.g., a path sequence). This direct construction forces the model to rely on internal planning rather than offloading logic to a code interpreter.
    
    \item \textbf{Improving (Refinement)}: Triggered when a valid solution is found. The LLM acts as an optimizer, retrieving \textit{historical context}—including past code/paths, performance metrics, and feedback trends. For ML, it performs \textbf{inductive reasoning} to adjust architecture or hyperparameters. For NP, it attempts to refine the discrete structure (e.g., shortening a TSP path) while maintaining validity constraints. This step tests the agent's ability to learn from the trajectory of past successes.
    
    \item \textbf{Debugging (Correction)}: Invoked when the environment returns a failure signal (e.g., runtime error or constraint violation). The agent analyzes the error logs provided by the environment to diagnose the fault. This step evaluates the agent's \textbf{self-correction} capability and its ability to recover from invalid states in the search space.
\end{itemize}
\section{Experiments and Results}
\label{sec:exp}
\begin{table*}[t]
\centering
\resizebox{\textwidth}{!}{
\tablestyle{6pt}{1.4}
\begin{tabular}{l|ccc|ccc|ccc}
\shline
\multirow{2}{*}{\bf Model} &  \multicolumn{3}{c|}{\bf 5 steps} &  \multicolumn{3}{c|}{\bf 10 steps} &  \multicolumn{3}{c}{\bf 20 steps}\\  \cline{2-10} 
 &  \textbf{Win Count} & \textbf {IR(w,w.o)}  & \textbf{EG} & \textbf{Win Count} & \textbf {IR(w,w.o)}  & \textbf{EG} & \textbf{Win Count} & \textbf {IR(w,w.o)}  & \textbf{EG} \\ \shline
 
\rowcolor{lightgray}
\multicolumn{10}{c}{\emph{Proprietary LLMs }}  \\ \shline
gpt-4o-2024-08-06 & 13/7 & 1.28 & 0.53 & 13/7 & 1.80 & 0.58 & \textbf{18/2} & 1.89 & 0.61 \\
gpt-4.1-2025-04-14 & 12/8 & 1.11 & 0.40 & 14/6 & 1.67 & 0.50 & 14/6 & \textbf{2.15} & 0.58 \\
gpt-o3-mini & 12/8 & 1.25 &\textbf{ 0.55} & 14/6 & 1.77 & \textbf{0.63} & 13/7 & 1.35 & \textbf{0.65} \\
gemini-2.0-flash & 13/7 & 1.08 & 0.45 & 14/6 & \textbf{2.08} & 0.48 & 14/6 & 1.95 & 0.53 \\
claude-3-5-sonnet-20241022 & 11/9 & 1.15 & 0.30 & 12/8 & 1.45 & 0.40 & 12/8 & 1.83 & 0.48 \\
claude-3-7-sonnet-20250219 & 14/6 & 1.00 & 0.43 & 14/6 & 1.11 & 0.48 & 14/6 & 1.35 & 0.53 \\
grok-3 & 12/8 & 1.16 & 0.50 & 14/6 & 1.36 & 0.55 & 15/5 & 1.29 & 0.63 \\
Deepseek-V3.1-Thinking & 12/8 & 1.13 & 0.50 & 12/8 & 1.18 & 0.58 & 11/9 & 1.03 & 0.60 \\
Qwen3-235B-Thinking & 12/8 & 1.18 & 0.52 & 13/7 & 1.13 & 0.55 & 14/6 & 1.13 & 0.62 \\
Qwen3-235B-Instruct  & 12/8 & 1.13 & 0.48 & 13/7 & 1.18 & 0.54 & 13/7 & 1.03 & 0.60 \\
\shline
\rowcolor{lightgray}
\multicolumn{10}{c}{\emph{Open-Source LLMs}}  \\ \shline
Internlm3-8b-instruct & 8/12 & 0.94 & 0.05 & 9/11 & 0.98 & 0.23 & 10/10 & 1.01 & 0.30 \\
Qwen2.5-3B-Instruct & 8/12 & 0.98 & 0.05 & 7/13 & 0.85 & 0.13 & 6/14 & 0.63 & 0.15 \\
Qwen2.5-7B-Instruct & 9/11 & 0.99 & 0.03 & 7/13 & 0.82 & 0.15 & 6/14 & 0.63 & 0.20 \\
Qwen2.5-14B-Instruct & 10/10 & 1.39 & 0.18 & 11/9 & 1.11 & 0.25 & 12/8 & 1.13 & 0.30 \\
Qwen2.5-32B-Instruct & 9/11 & 1.38 & 0.25 & 10/10 & 1.21 & 0.33 & 11/9 & 1.19 & 0.45 \\
Qwen2.5-72B-Instruct & \textbf{15/5} & \textbf{1.58} & 0.40 & \textbf{15/5} & 1.47 & 0.43 & 15/5 & 1.61 & 0.45 \\
Qwen3-8B & 9/11 & 0.99 & 0.28 & 10/10 & 1.24 & 0.33 & 12/8 & 1.16 & 0.38 \\
Qwen3-30B-A3B & 10/10 & 1.06 & 0.45 & 11/9 & 1.08 & 0.53 & 12/8 & 1.14 & 0.54 \\
Qwen3-32B & 10/10 & 1.07 & 0.18 & 11/9 & 1.19 & 0.48 & 12/8 & 1.28 & 0.58 \\
\shline
\end{tabular}
}
\caption{Evaluation Results of LLMs on OPT-BENCH-ML, comparing both closed-source and open-source models, including general and reasoning models.}
\label{tab:main-results}
\vspace{-1.5em}
\end{table*}
\subsection{Experimental Setup}
\label{sec:exp_env}
\textbf{Environments.} OPT-BENCH-ML experiments utilized a standard CPU environment (4 cores, 32GB RAM), while OPT-BENCH-NP required minimal resources (2 cores, 16GB RAM). We accessed proprietary models via API and open-source models (3B--72B) via LMDeploy.

\noindent\textbf{Baselines and Metrics.} To rigorous quantify the \textbf{Self-Optimization} capability—defined as the ability to improve over time using environmental feedback—we establish the following evaluation protocols:

\subsubsection{Evaluation Metrics and Baselines}
To rigorously quantify the \textbf{Self-Optimization} capability—defined as the agent's ability to evolve from a crude initial state to an expert-level solution using environmental feedback—we establish the following evaluation protocols:

\begin{itemize}
    \item \textbf{Baselines (The Optimization Bounds):}
    \begin{enumerate}
        \item \textit{Lower Bound (Random Rollout):} The performance of the agent generating solutions independently without historical context (memory-less). This represents "blind" trials and serves as the baseline for assessing whether the agent is truly learning or merely guessing.
        \item \textit{Upper Bound (Human Expert):} The Gold Medal solution (for ML) or heuristic optimal (for NP), representing the functional ceiling of task performance.
    \end{enumerate}

\item \textbf{Win Count (Feedback Utility):}
Defined as the number of tasks in which the OPT-Agent outperforms the \textit{Random Rollout} baseline. This metric quantifies the model’s ability to perform {self‑optimization}. A high Win Count indicates that the model is successfully interpreting environmental feedback and using it to guide performance improvement.

\item \textbf{Improvement Rate (IR):}
A quantitative measure of \textbf{learning efficiency}, primarily used to capture relative performance gains in ML tasks when compared with the Random Rollout baseline. It is defined as:
\[
\mathrm{IR}(\alpha,\beta)
= \frac{1}{n} \sum_{i=1}^n \frac{\alpha_i}{\beta_i},
\]
where $\alpha_i$ denotes the metric value after optimization, and $\beta_i$ denotes the corresponding baseline value (either the initial solution or a memory‑less generation). This metric reflects the \textit{magnitude of progress} achieved along the optimization trajectory; values of $\mathrm{IR}>1.0$ indicate that the LLM is {effectively leveraging environmental feedback} to improve performance.
    \item \textbf{Expert Gap (EG):} 
    To address the scale heterogeneity across 30 diverse tasks (e.g., MSE vs. Accuracy), we compute the \textit{Normalized Gap} relative to the human expert. For a task with metric $M$, the score is normalized as:
    \[
    \mathrm{Score}_{norm} = \frac{M_{current} - M_{initial}}{M_{expert} - M_{initial}}
    \]
    This serves as the definitive measure of \textbf{solution quality}. It quantifies the extent to which the agent's self-optimization loop bridges the gap between the initial 'Draft' and the 'Expert' solution, enabling fair comparisons across diverse domains.

    \item \textbf{Buggy Rate:} 
The proportion of invalid solution attempts (e.g., syntax errors or constraint violations). A decreasing buggy rate over iterations indicates that the agent is not only optimizing for task performance but also progressively learning to satisfy the environment’s structural constraints—particularly in NP tasks.
\end{itemize}

\subsection{Main Results: Dynamics of Self-Optimization}
\label{exp:main_exp}

\textbf{Continuous Parametric Optimization (ML).} 
Table~\ref{tab:main-results} validates our hypothesis: in continuous search spaces, strong LLMs effectively function as \textit{inductive optimizers}, interpreting numerical feedback as directional gradients to guide self-optimization. Proprietary models (e.g., \texttt{gpt-4o}) achieve a dominant {18/2 Win Count} over the random baseline at 20 steps. This confirms that performance gains stem from purposeful, \textbf{history-driven refinement}—effectively pruning the search space based on accumulated environmental signals—rather than from stochastic fluctuations.

\noindent\textbf{Scaling Law of Model Size.} 
We observe a sharp divergence in performance based on model scale (e.g., Expert Gap {0.45} for \texttt{Qwen2.5-72B} vs.\ {0.20} for \texttt{7B}). This suggests a potential \textbf{cognitive threshold}: smaller models tend to perceive complex error traces as noise due to limited semantic working memory, whereas larger models exhibit emergent capabilities for {multi-step inductive reasoning}.

\noindent\textbf{Reasoning vs.\ General Models.} 
Reasoning models (e.g., \texttt{gpt-o3-mini}) consistently outperform general instruction-tuned counterparts, achieving the highest Expert Gap closure ({0.65}). This indicates that the \textit{Chain-of-Thought (CoT)} paradigm is critical for \textit{fine-grained optimization}, enabling agents to deduce the causal mechanism of failure rather than resorting to random parameter guessing.

\noindent\textbf{Optimization Horizon.} 
Extending the optimization trajectory to 20 steps yields substantial gains for proprietary models (e.g., \texttt{gpt-4.1} achieves an IR of 2.15), while weaker models plateau early. This highlights that \textbf{long-horizon self-optimization} requires strong contextual retention: without it, agents lose focus and regress to near-random exploration.

\textbf{Discrete Combinatorial Domains (NP).} 
Table~\ref{tab:main-results-np} presents a contrasting landscape: in combinatorial search spaces, environmental feedback yields diminishing returns for self-optimization.

\noindent\textbf{The Feedback Efficiency Paradox.} 
Unlike the dominance observed in ML tasks, Win Counts here are more balanced (e.g., \texttt{gpt-4o}: 4/6), revealing a \textbf{feedback misinterpretation} phenomenon. Discrete error signals (e.g., ``Invalid Cycle'') lack directional gradients, making it difficult for standard models to translate them into structural repairs. As a result, self-optimization often degenerates into ``random search with memory.''

\noindent\textbf{Reasoning vs General Models.} 
Reasoning models (e.g., \texttt{Deepseek-V3.1-Thinking}) achieve a {0.00 buggy rate} and the highest Expert Gap ($\sim$0.79), confirming that {Chain-of-Thought (CoT)} reasoning is \textit{essential} for maintaining global topological consistency. These models can verify solution validity in ways that instruction-tuned models often fail to do.

\noindent\textbf{The Validity Bottleneck.} 
Smaller open-source models encounter a \textbf{feasibility threshold} (buggy rates $\sim$0.80). Unlike soft failure modes in ML domains, NP tasks impose a hard ``valid/invalid'' constraint. Failure to cross this threshold renders historical feedback ineffective, as the agent never establishes a valid baseline from which to optimize.

\noindent\textbf{Optimization Horizon and the Deductive Ceiling.} 
Larger general models (e.g., \texttt{Qwen2.5-72B}) exhibit moderate gains (EG $0.33 \rightarrow 0.47$) over 20 steps, while smaller models plateau early due to insufficient planning depth for complex topological transformations. Interestingly, reasoning models also show limited improvement despite strong initial performance, possibly due to \textbf{reinforcement learning alignment constraints} that inhibit long-horizon exploration.

\vspace{-1em}
\begin{table*}[t]
\centering
\small 
\resizebox{\textwidth}{!}{
\tablestyle{4pt}{1.4}
\begin{tabular}{l|c|cc|cc|c|cc|cc|c|cc|cc}
\shline
\multirow{3}{*}{\bf Model} &  \multicolumn{5}{c|}{\bf 5 steps} &  \multicolumn{5}{c|}{\bf 10 steps} &  \multicolumn{5}{c}{\bf 20 steps}\\  \cline{2-16} 
 &  \multirow{2}{*}{\textbf{Win Count}} & \multicolumn{2}{c|}{\textbf {Buggy Rate}}  & \multicolumn{2}{c|}{\textbf{EG}} & \multirow{2}{*}{\textbf{Win Count}} & \multicolumn{2}{c|}{\textbf {Buggy Rate}}  & \multicolumn{2}{c|}{\textbf{EG}} & \multirow{2}{*}{\textbf{Win Count}} & \multicolumn{2}{c|}{\textbf {Buggy Rate}}  & \multicolumn{2}{c}{\textbf{EG}} \\ 
 \cline{3-6}\cline{8-11}\cline{13-16}
 & & w & w.o & w & w.o & & w & w.o & w & w.o & & w & w.o & w & w.o\\\shline
 
\rowcolor{lightgray}
\multicolumn{16}{c}{\emph{Proprietary LLMs }}  \\ \shline
gpt-4o-2024-08-06 & 5/5 & 0.30 & 0.34 & 0.41 & 0.40 & 4/6 & 0.26 & 0.30 & 0.44 & 0.44 & 4/6 & 0.22 & 0.18 & 0.47 & 0.54 \\
gpt-4.1-2025-04-14 & 4/6 & 0.10 & 0.10 & 0.45 & 0.47 & 4/6 & 0.08 & 0.08 & 0.47 & 0.48 & 4/6 & 0.04 & 0.02 & 0.53 & 0.55 \\
gpt-o3-mini & 5/5 & \textbf{0.00} & \textbf{0.00} & 0.72 & 0.68 & 6/4 & \textbf{0.00} & \textbf{0.00} & 0.72 & 0.69 & 5/5 & \textbf{0.00} & 0.00 & 0.73 & 0.69 \\
gemini-2.0-flash & 5/5 & 0.16 & 0.20 & 0.40 & 0.42 & 6/4 & 0.06 & 0.12 & 0.44 & 0.43 & \textbf{7/3} & 0.06 & 0.10 & 0.45 & 0.40 \\
claude-3-5-sonnet-20241022 & 7/3 & 0.20 & 0.32 & 0.49 & 0.44 & 7/3 & 0.18 & 0.30 & 0.51 & 0.46 & \textbf{7/3} & 0.18 & 0.28 & 0.52 & 0.47 \\
claude-3-7-sonnet-20250219 & \textbf{8/2} & 0.10 & 0.24 & 0.62 & 0.52 & \textbf{8/2} & 0.08 & 0.22 & 0.64 & 0.55 & \textbf{7/3} & 0.06 & 0.12 & 0.66 & 0.58 \\
grok-3 & 5/5 & 0.16 & 0.14 & 0.52 & 0.55 & 5/5 & {0.14} & {0.12} & 0.57 & 0.56 & 6/4 & {0.12} & {0.11} & 0.62 & 0.58 \\
Deepseek-V3.1-Thinking & 6/4 & \textbf{0.00} & \textbf{0.00} & \textbf{0.76} &{ 0.74} & 5/5 & \textbf{0.00} & \textbf{0.00} & \textbf{0.78} & {0.75} & 5/5 & 0.00 & 0.00 & \textbf{0.79} &\textbf{ 0.76} \\
Qwen3-235B-Thinking & 6/4 & \textbf{0.00} & \textbf{0.00} & {0.74} &\textbf{ 0.76} & 5/5 & \textbf{0.00} & \textbf{0.00} & {0.75} & \textbf{0.76} & 5/5 & 0.00 & 0.00 & \textbf{0.76} &\textbf{ 0.76} \\
Qwen3-235B-Instruct & 6/4 & 0.16 & 0.15 & 0.48 & 0.50 & 6/4 & 0.10 & 0.10 & 0.53 & 0.52 & 6/4 & 0.06 & 0.10 & 0.57 & 0.55 \\
\shline
\rowcolor{lightgray}
\multicolumn{16}{c}{\emph{Open-Source LLMs}} \\
\shline
Internlm3-8b-Instruct & 4/6 & 0.52 & 0.54 & 0.12 & 0.14 & 6/4 & 0.30 & 0.42 & 0.20 & 0.18 & 6/4 & 0.18 & 0.20 & 0.24 & 0.20 \\
Qwen2.5-3B-Instruct & 6/4 & 0.64 & 0.80 & 0.12 & 0.03 & 7/3 & 0.46 & 0.68 & 0.18 & 0.09 & 8/2 & 0.38 & 0.58 & 0.21 & 0.14 \\
Qwen2.5-7B-Instruct & 6/4 & 0.54 & 0.70 & 0.22 & 0.14 & 7/3 & 0.38 & 0.54 & 0.31 & 0.23 & 6/4 & 0.24 & 0.28 & 0.35 & 0.33 \\
Qwen2.5-14B-Instruct & 5/5 & 0.40 & 0.60 & 0.31 & 0.22 & 5/5 & 0.24 & 0.28 & 0.41 & 0.39 & 6/4 & 0.18 & 0.18 & 0.47 & 0.43 \\
Qwen2.5-32B-Instruct & 5/5 & 0.42 & 0.44 & 0.33 & 0.35 & 5/5 & 0.30 & 0.28 & 0.41 & 0.43 & 6/4 & 0.20 & 0.20 & 0.52 & 0.46 \\
Qwen2.5-72B-Instruct & 5/5 & 0.40 & 0.50 & 0.33 & 0.30 & 5/5 & 0.32 & 0.38 & 0.39 & 0.38 & 4/6 & 0.22 & 0.32 & 0.47 & 0.44 \\
Qwen3-8B & 5/5 & 0.28 & 0.22 & 0.51 & 0.56 & 5/5 & 0.20 & 0.20 & 0.57 & 0.58 & 5/5 & 0.12 & 0.20 & 0.63 & 0.60 \\
Qwen3-30B-A3B & 5/5 & 0.04 & 0.12 & 0.60 & 0.59 & 5/5 & \textbf{0.00} & 0.06 & 0.65 & 0.65 & 6/4 &\textbf{ 0.00} & \textbf{0.00} & 0.69 & 0.68 \\
Qwen3-32B & 5/5 & 0.14 & 0.12 & 0.52 & 0.57 & 4/6 & 0.10 & 0.10 & 0.57 & 0.59 & 5/5 & 0.06 & 0.10 & 0.60 & 0.60 \\
\shline
\end{tabular}
}
\caption{Evaluation Results of LLMs on OPT-BENCH-ML, comparing both closed-source and open-source models, including general and reasoning models.}
\label{tab:main-results-np}
\end{table*}

\subsection{Ablation Study: Stability vs. Exploration}
\label{exp:ablation}
To decouple the effects of sampling randomness from intrinsic reasoning, we investigate how Temperature ($T$) modulates the \textbf{Self-Optimization} trajectory (Tables 3 and 4).

\begin{itemize}
    \item \textbf{Continuous Domains (ML):} 
    For strong models like \texttt{gpt-4o}, performance peaks at $T=0$ (13/7 Win Count) and degrades significantly at $T=0.8$ (10/10). 
    \textit{Insight:} In continuous spaces, environmental feedback acts as a specific directional gradient (e.g., "decrease learning rate"). Setting $T=0$ ensures the agent strictly adheres to this signal, maximizing \textbf{Exploitation}. Higher temperatures introduce "adversarial noise," disrupting the precise numerical fine-tuning required for convergence and effectively diluting the value of historical context.

    \item \textbf{Discrete Domains (NP):} 
    The dynamics shift in combinatorial spaces. While $T=0$ ensures consistency, a moderate temperature ($T=0.2$) often yields lower Buggy Rates for models like \texttt{gpt-4o} (0.18 vs. 0.28 at $T=0$).
    \textit{Insight:} Unlike the smooth landscape of ML, NP problems are characterized by "rugged" local optima. Deterministic decoding can cause the agent to fixate on a specific invalid structural pattern. Slight stochasticity ($T=0.2$) provides the necessary \textbf{Exploration} to escape these local traps. However, excessive randomness ($T=0.8$) shatters the logical coherence required to maintain constraints, causing Win Counts to drop (e.g., \texttt{gpt-4o} falls to 4/6) as the optimization degenerates into random guessing.
\end{itemize}

\begin{table*}[t]
\centering
\small 
\resizebox{\textwidth}{!}{
\tablestyle{6pt}{1.4}
\begin{tabular}{l|ccc|ccc|ccc}
\shline
\multirow{2}{*}{\bf Model} &  \multicolumn{3}{c|}{\bf Temperature=0} &  \multicolumn{3}{c|}{\bf Temperature=0.2} &  \multicolumn{3}{c}{\bf Temperature=0.8}\\  \cline{2-10} 
 &  \textbf{Win Count} & \textbf {IR(w,w.o)}  & \textbf{EG} & \textbf{Win Count} & \textbf {IR(w,w.o)}  & \textbf{EG} & \textbf{Win Count} & \textbf {IR(w,w.o)}  & \textbf{EG} \\ 
\shline
gpt-4o-2024-08-06 & \textbf{13/7} & \textbf{1.58} & 0.58 & 9/11 & 1.19 & 0.50 & 10/10 & 1.10 & 0.40 \\
grok-3 & 9/11 & 1.01 & 0.65 & \textbf{11/9} & \textbf{1.29} & 0.45 & 8/12 & 1.04 & 0.43 \\
Qwen2.5-72B-Instruct & 10/10 & 1.05 & 0.43 & 11/9 & 1.04 & 0.40 & \textbf{11/9} & \textbf{1.11} & 0.44 \\
\shline
\end{tabular}
}
\caption{{Evaluation Results of LLMs on OPT-BENCH-ML Across Different Temperature Settings.}}
\label{tab:ablation_temp-ml}
\vspace{-1em}
\end{table*}
\begin{table*}[t]
\centering
\small 
\resizebox{\textwidth}{!}{
\tablestyle{4pt}{1.4}
\begin{tabular}{l|c|cc|cc|c|cc|cc|c|cc|cc}
\shline
\multirow{3}{*}{\bf Model} &  \multicolumn{5}{c|}{\bf Temperature=0} &  \multicolumn{5}{c|}{\bf Temperature=0.2} &  \multicolumn{5}{c}{\bf Temperature=0.8}\\  \cline{2-16} 
 &  \multirow{2}{*}{\textbf{Win Count}} & \multicolumn{2}{c|}{\textbf {Buggy Rate}}  & \multicolumn{2}{c|}{\textbf{EG}} & \multirow{2}{*}{\textbf{Win Count}} & \multicolumn{2}{c|}{\textbf {Buggy Rate}}  & \multicolumn{2}{c|}{\textbf{EG}} & \multirow{2}{*}{\textbf{Win Count}} & \multicolumn{2}{c|}{\textbf {Buggy Rate}}  & \multicolumn{2}{c}{\textbf{EG}} \\ 
 \cline{3-6}\cline{8-11}\cline{13-16}
 & & w & w.o & w& w.o& & w& w.o & w& w.o& & w& w.o& w& w.o\\\shline
 
gpt-4o-2024-08-06 & \textbf{5/5} & 0.28&0.24 & 0.48& 0.44 & \textbf{5/5} &{0.18}&0.28 & {0.51}& 0.47 & 4/6 & {0.18}&0.22 & 0.47&{0.49} \\
grok-3 &\textbf{5/5} & \textbf{0.14}&\textbf{0.18} & \textbf{0.54}&\textbf{0.52} & \textbf{5/5} & \textbf{0.12}&\textbf{0.14} & \textbf{0.56}&\textbf{0.53} & 4/6 & \textbf{0.00} &\textbf{0.00} & \textbf{0.60}&\textbf{0.61} \\
Qwen2.5-72B-Instruct & \textbf{5/5} & 0.30&0.34 & 0.47&0.46 & 4/6 & 0.24&0.26 & 0.44 & 0.46 & \textbf{5/5} & 0.25&0.26 & 0.46& 0.44 \\\shline
\end{tabular}

}
\caption{Evaluation Results of LLMs on OPT-BENCH-NP across Different Temperature Settings.}
\label{tab:ablation_temp-np}
\vspace{-1.5em}
\end{table*}

\subsection{Discussion: The Limits of Intrinsic Self Optimization}
\label{exp:analysis_np_ml}

Our findings highlight a critical boundary in LLM self-optimization cognition. 
\noindent \textbf{Signal Interpretation Divergence:} Self-optimization hinges on accurately mapping \textit{Environmental Feedback} to \textit{Action Updates}. We observe that LLMs excel in ML domains where feedback is semantic or numerical, facilitating coherent, incremental refinement. Conversely, they falter in NP tasks due to the challenge of maintaining complex structural dependencies. For instance, in the Hamiltonian cycle problem, while a human would locally repair a disconnected edge, LLMs often discard the entire history to generate a completely new solution, as shown in Figure~\ref{fig:agent-trace}. This inability to perform \textit{incremental structural editing} explains why historical feedback yields diminishing returns in discrete combinatorial spaces compared to continuous ML landscapes.

\noindent \textbf{The Human Gap:} Even with self-optimization, a significant gap persists compared to the Human Expert Upper Bound. While agents can improve over their initial drafts, they rarely reach the global optima that humans find through principled heuristic derivation, suggesting that current "Self-Optimization" is essentially a \textit{local search} mechanism bounded by the model's inherent reasoning depth.
\section{Related Work}
\paragraph{LLM Evaluation}
Early benchmarks like MMLU~\cite{hendrycks2020measuring} and BIG-bench~\cite{srivastava2022beyond} assessed broad knowledge but remained confined to static, multiple-choice formats. While subsequent reasoning-focused evaluations—ranging from math (MATH~\cite{hendrycks2021measuring}, GSM8K~\cite{cobbe2021training}) and code (HumanEval~\cite{chen2021evaluating}, MBPP~\cite{austin2021program}) to combinatorial problems (NP-Engine~\cite{li2025npengine})—have advanced the field, they predominantly rely on \textbf{one-shot, open-loop paradigms}. Even with Chain-of-Thought prompting~\cite{wei2022chain}, these benchmarks evaluate \textit{instantaneous deduction} rather than \textit{adaptive learning}. Algotune~\cite{algotune} and ALEBench~\cite{alebench} emphasize “algorithm engineering”—generating code that is correct and efficient (e.g., runtime/complexity). Notably, NPHardEval~\cite{fan2023nphardeval} introduces complexity classes but remains a single-pass solvability test, failing to capture the \textbf{iterative self-correction} dynamics essential for autonomous optimization.

\paragraph{LLM Agents}
Recent frameworks have empowered LLMs with tool use (ReAct~\cite{yao2022react}, RDAgent~\cite{RDagent}, Toolformer~\cite{schick2023toolformer}) and feedback loops (Reflexion~\cite{shinn2023reflexion}). However, existing agent benchmarks primarily focus on \textbf{task completion} rather than \textbf{solution refinement}. Benchmarks like AgentBench~\cite{liu2023agentbench}, WebShop~\cite{yao2022webshop}, and WebArena~\cite{zhou2023webarena} evaluate whether an agent can \textit{execute} a sequence of actions to finish a task, not whether it can \textit{optimize} a metric over time. MLE-Bench~\cite{openai2024mlebench} targets machine learning engineering but emphasizes code execution success over the cognitive trajectory of hyperparameter tuning. Similarly, IOLBench~\cite{zhang2024iolbench} focuses on linguistic reasoning.
A critical gap remains for a unified benchmark that strictly evaluates \textbf{Intrinsic Self-Optimization}—the ability to leverage environmental feedback to climb performance landscapes—across both continuous (ML) and discrete (NP) domains. OPT-BENCH addresses this by shifting focus from "Can the agent run the code?" to "Can the agent evolve the solution?"

\vspace{-0.5em}
\section{Conclusion}
\vspace{-0.5em}
We introduced \textbf{OPT-BENCH} and \textbf{OPT-Agent} to evaluate the ability of LLMs to perform \textit{iterative self-optimization} across contrasting search spaces. By juxtaposing 20 continuous (ML) and 10 discrete (NP) environments, our evaluation of 19 LLMs across 7 model families reveals a clear divergence: while models exhibit strong performance on \textbf{Continuous Parametric Optimization}—effectively leveraging numerical feedback for iterative refinement—they struggle with \textbf{Combinatorial Reasoning}, often failing to translate discrete error signals into structural repairs.
We further find that reasoning-enhanced models consistently outperform general models, particularly in NP domains where surpassing the feasibility threshold requires structured reasoning. Such models are essential for handling combinatorial constraints. However, the persistent gap between even the strongest LLMs and \textbf{Human Expert Baselines} reveals a fundamental limitation: achieving true autonomous optimization demands more than local pattern matching—it requires global planning and high-level reasoning.
\section*{Limitations}
Due to resource constraints, we were unable to include additional state-of-the-art models such as the Gemini 3 series, OpenAI GPT-5.2, and Claude 4.5 in our experiments. In addition, our current benchmark contains only 30 environments. In future work, we plan to scale the benchmark to a larger and more diverse set of environments in order to enable more comprehensive and robust evaluation.
\bibliography{custom}

\appendix

\newpage
\appendix
\section{Appendix}

\subsection{Use of Large Language Models}
Large Language Models are used for grammar check and polishing in this paper.

\subsection{Ethics Statement.}
All datasets used in this study are obtained from public sources and are freely available for academic research. Therefore, we do not anticipate that the data used in this work pose any significant privacy risks.

\subsection{Draft Setting}
Unlike the refine setting, the draft setting requires models to generate solutions from scratch, providing a more direct test of their optimization capabilities. Results for three strong models are shown in Table~\ref{tab:ablation_draft}. Notably, the open-source \texttt{Qwen2.5-72B-Instruct} consistently shows higher buggy rates than proprietary models, reflecting greater difficulty in producing valid solutions during draft optimization. However, except for \texttt{grok-3} at 5 steps, all models achieve higher improvement rates (\textit{IR(d,r)}) than in the refine setting, indicating that draft optimization can outperform traditional refinement when valid solutions are found. The \textit{Win Count} metric further shows that using historical information during draft optimization improves performance across all step counts. Increasing the number of steps yields significant gains in both improvement rates and win counts, highlighting the value of iterative refinement. These results emphasize the importance of managing the trade-off between exploration and solution validity in draft optimization and suggest that reducing buggy rates is key to advancing both proprietary and open-source LLMs in this setting.
\begin{table*}[ht]
\centering
\small 
\resizebox{\textwidth}{!}{
\tablestyle{6pt}{1.4}
\begin{tabular}{l|cc|c|c|cc|c|c|cc|c|c}
\shline
\multirow{3}{*}{\bf Model} &  \multicolumn{4}{c|}{\bf 5 steps} &  \multicolumn{4}{c|}{\bf 10 steps} &  \multicolumn{4}{c}{\bf 20 steps}\\  \cline{2-13} 
 &  \multicolumn{2}{c|}{\bf{Buggy Rate}} & \multirow{2}{*}{\textbf {Win Count}}  & \multirow{2}{*}{\textbf{IR(d,r) }} &  \multicolumn{2}{c|}{\bf{Buggy Rate}} & \multirow{2}{*}{\textbf {Win Count}}  & \multirow{2}{*}{\textbf{IR(d,r) }} &  \multicolumn{2}{c|}{\bf{Buggy Rate}} & \multirow{2}{*}{\textbf {Win Count}}  & \multirow{2}{*}{\textbf{IR(d,r) }} \\ \cline{2-3} \cline{6-7} \cline{10-11} 
 & w & w.o & & &w & w.o & & & w & w.o & & \\ \shline
 
gpt-4o-2024-08-06 & 0.20 & 0.20 & 8/6 & 1.24 & \textbf{0.15} & \textbf{0.15} & \textbf{10/5} & 1.38 & \textbf{0.15} & \textbf{0.15} & \textbf{11/4} & 1.41 \\
grok-3 & \textbf{0.15} & \textbf{0.15} & \textbf{10/5} & 0.80 & \textbf{0.15} & \textbf{0.15} & \textbf{10/5} & 1.38 & 0.15 & 0.15 & \textbf{11/4} & \textbf{1.42} \\
Qwen2.5-72B-Instruct & 0.40 & 0.35 & 5/5 & \textbf{1.54} & 0.30 & 0.30 & 6/6 & \textbf{1.94} & 0.20 & 0.20 & 8/7 & 1.12 \\
\shline
\end{tabular}
}
\caption{\textbf{Evaluation Results of LLMs under Draft Settings.} Metrics include \textit{Buggy Rate}, denoting the proportion of invalid solutions; \textit{Win Count}, comparing OPT-Agent-draft optimization against the baseline without historical information; and \textit{IR(d,r)}, the improvement rate comparing OPT-Agent-draft optimization to OPT-Agent-refine.}
\label{tab:ablation_draft}

\end{table*}

\subsection{OPT-BENCH Dataset}
\label{sec:task_list}
The detailed information regarding the 20 Machine Learning (ML) tasks and 10 NP problems used in our OPT-BENCH is comprehensively summarized in Table~\ref{table:app_20ML_coms} and Table~\ref{table:app_10NP_coms}, respectively. These tables provide concise descriptions of each task or problem, along with their corresponding evaluation metrics, which serve as the foundation for assessing the performance of OPT-Agent across diverse optimization scenarios.

\begin{table*}[ht]
\centering
\resizebox{\textwidth}{!}{
\tablestyle{6pt}{1.4}
\begin{tabular}{lll}
\hline
\textbf{Kaggle Competition} & \textbf{Description} & \textbf{Metric} \\
\hline

\href{https://www.kaggle.com/c/bike-sharing-demand/}{bike-sharing-demand} & Forecast use of a city bikeshare system & Root Mean Squared Logarithmic Error (RMSLE) \textdownarrow \\

\href{https://www.kaggle.com/c/competitive-data-science-predict-future-sales}{competitive-data-science-predict-future-sales} & Predict total sales for every product and store & Root Mean Squared Error (RMSE) \textdownarrow \\

\href{https://www.kaggle.com/c/house-prices-advanced-regression-techniques/}{house-prices-advanced-regression-techniques} & Predict house sales prices & Root-Mean-Squared-Error (RMSE) \textdownarrow \\

\href{https://www.kaggle.com/competitions/london-house-price-prediction-advanced-techniques}{london-house-price-prediction-advanced-techniques} & Predict London house prices & Mean Absolute Error (MAE) \textdownarrow \\

\href{http://kaggle.com/c/playground-series-s3e14}{playground-series-s3e14} & Predicting wild blueberry yields & Mean Absolute Error (MAE) \textdownarrow \\

\href{https://www.kaggle.com/c/playground-series-s3e16/}{playground-series-s3e16} & Predict the age of crabs & Mean Absolute Error (MAE) \textdownarrow \\

\href{https://www.kaggle.com/c/playground-series-s3e19/}{playground-series-s3e19} & Forecast Mini-Course Sales & Symmetric Mean Absolute Percentage Error (SMAPE) \textdownarrow \\

\href{https://www.kaggle.com/c/playground-series-s3e22/}{playground-series-s3e22} & Predict Health Outcomes of Horses & micro-averaged F1-Score \textuparrow \\

\href{https://www.kaggle.com/c/playground-series-s3e24}{playground-series-s3e24} & Binary Prediction of Smoker Status using Bio-Signals & area under the ROC curve \textuparrow \\

\href{https://www.kaggle.com/c/playground-series-s3e25}{playground-series-s3e25} & Regression with a Mohs Hardness Dataset & Median Absolute Error (MedAE) \textdownarrow \\

\href{https://www.kaggle.com/competitions/playground-series-s3e3}{playground-series-s3e3} & Binary Classification with a Tabular Employee Attrition Dataset & area under the ROC curve \textuparrow \\

\href{https://www.kaggle.com/competitions/playground-series-s3e5}{playground-series-s3e5} & Ordinal Regression with a Tabular Wine Quality Dataset & quadratic weighted kappa \textuparrow \\

\href{https://www.kaggle.com/competitions/playground-series-s4e2}{playground-series-s4e2} & Multi-Class Prediction of Obesity Risk & Accuracy \textuparrow \\

\href{https://www.kaggle.com/competitions/sentiment-analysis-on-movie-reviews/}{sentiment-analysis-on-movie-reviews} & Classify the sentiment of sentences from the Rotten Tomatoes dataset & classification accuracy \textuparrow \\

\href{https://www.kaggle.com/c/spaceship-titanic}{spaceship-titanic} & Predict which passengers are transported to an alternate dimension & classification accuracy \textuparrow \\

\href{https://www.kaggle.com/c/tabular-playground-series-aug-2022}{tabular-playground-series-aug-2022} & Predict product failures & area under the ROC curve \textuparrow \\

\href{https://www.kaggle.com/c/tabular-playground-series-feb-2021}{tabular-playground-series-feb-2021} & Predict the amount of an insurance claim & Root-Mean-Squared-Error (RMSE) \textdownarrow \\

\href{https://www.kaggle.com/c/tabular-playground-series-jul-2021}{tabular-playground-series-jul-2021} & Predict air pollution in a city  & Root Mean Squared Logarithmic Error (RMSLE) \textdownarrow \\

\href{https://www.kaggle.com/c/tabular-playground-series-sep-2022}{tabular-playground-series-sep-2022} & Predict book sales & Symmetric Mean Absolute Percentage Error (SMAPE) \textdownarrow \\

\href{https://www.kaggle.com/competitions/telstra-recruiting-network}{telstra-recruiting-network} & Predict the severity of service disruptions on their network & multi-class logarithmic loss \textdownarrow \\
\hline
\end{tabular} 
}
\caption{\textbf{Kaggle Machine Learning Competitions with Description and Metric.}}
\label{table:app_20ML_coms}
\end{table*}

\begin{table*}[ht]
\centering
\resizebox{\textwidth}{!}{
\tablestyle{6pt}{1.4}
\begin{tabular}{llll}
\hline
\textbf{NP Problem} & \textbf{Description} & \textbf{Metric} & \textbf{Human Expert Baseline} \\
\hline
Graph Coloring Problem (GCP) & Use the minimum number of colors necessary to achieve a valid coloring & Color Number $\downarrow$ & Heuristic Search \\

Hamiltonian Cycle & Find the largest possible valid Hamiltonian circuit & Path Length $\uparrow$ & Heuristic Search \\

Knapsack & Choose a subset of items to pack into a limited-capacity bag & Total Item Weight $\uparrow$ & Greedy Heuristic \\

Maximum Clique Problem & Find the largest clique (a subset of vertices all connected to each other) in a given graph & Clique Size $\uparrow$ & Simulated Annealing \\

Maximum Set & Find the largest subset of a set under constraints & Set Size $\uparrow$ & Greedy Heuristic \\

Meeting Schedule & Schedule meetings for as many participants as possible under various constraints & Total Attendees $\uparrow$ & Heuristic Scheduling \\

Minimum Cut & Find the minimum cut in a graph & Cut Weight $\downarrow$ & Graph Heuristic \\

Set Cover & Select a minimal number of sets from a collection such that their union covers all elements of a universal set & Subset Number $\downarrow$ & Greedy Heuristic \\

Subset Sum & Find a subset of a set of numbers that adds up to a specific target value & Indice Number $\uparrow$ & Heuristic Search \\

Traveling Salesman Problem (TSP) & Find the shortest possible route that visits each city once and returns to the starting point & Route Length $\downarrow$ & Simulated Annealing \\
\hline
\end{tabular}
}
\vspace{2mm}
\caption{\textbf{NP problems, task descriptions, evaluation metrics, and human expert baselines.} The Human Expert Baseline corresponds to standard heuristic algorithms.}
\label{table:app_10NP_coms}
\end{table*}

\begin{table*}[ht]
\centering
\resizebox{\textwidth}{!}{
\tablestyle{6pt}{1.4}
\begin{tabular}{llll}
\hline
\textbf{Feature} & \textbf{SWEBench/ AlgoTune} & \textbf{MLE-Bench} & \textbf{OPT-BENCH (Ours)} \\
\hline
{Core Goal} & Bug fixing / Code efficiency & End-to-end engineering & Iterative feedback utilization \\
{Feedback Signal} & Unit tests (pass/fail) & Leaderboard score & Scalar metric + error traces (ML) / objective + constraint checks (NP) \\
{Search Space} & Syntax/logic correctness & Broad workflow & Continuous (ML) vs.\ discrete (NP) \\
{Cognitive Focus} & Coding capability & Agentic workflow & Self-optimization \& correction \\
\hline
\end{tabular}
}
\vspace{2mm}
\caption{\textbf{Comparison of representative benchmarks.} We compare SWEBench/AlgoTune, MLE-Bench, and our OPT-BENCH in terms of core goal, feedback signal, search space, and cognitive focus. The OPT-BENCH emphasizes iterative feedback utilization and self-optimization under both continuous (ML) and discrete (NP) settings.}
\label{table:bench}
\end{table*}

\section{OPT-Agent Prompt}

In this section, as illustrated in Figure~\ref{fig:agent_ml_np} and Figure~\ref{fig:prompt_ml_np}, we provide a comprehensive overview of the prompt templates used in both OPT-Agent-ML and OPT-Agent-NP. These prompts guide the model through three distinct types of actions—draft, improve, and debug—by delivering task-specific context and structured response formats. Specifically, the OPT-Agent-ML prompts focus on instructing the model for machine learning tasks, while the OPT-Agent-NP prompts are carefully designed to include structured task descriptions, input-output examples, and response formatting guidelines, enabling the model to systematically address and refine solutions to NP problems.

\begin{figure*}[ht]
    \centering
    
    \begin{minipage}{\textwidth}
        \centering
        \begin{AIbox}{OPT-Agent-ML}
        {   
            \textbf{\textcolor{orange}{Introduction (draft): }}
            You are a Kaggle grandmaster attending a competition \textcolor{brown}{\textless task type\textgreater}. In order to win this competition, you need to come up with an exceptional and creative plan. To address this problem, I will provide you with the specific task description, the evaluation metrics to be used, training set and submission format in sequence.

            \textbf{\textcolor{green}{Introduction (improve): }}
            You are a Kaggle grandmaster attending a competition \textcolor{brown}{\textless task type\textgreater}. You have been provided with previously developed solution, and your task is to improve it in order to increase the performance in test dataset. Review previous solution and improve based on it. You can only modify the model, optimizer, or hyperparameters, and adjust feature engineering for compatibility.

            \textbf{\textcolor{purple}{Introduction (draft): }}
            You are a Kaggle grandmaster attending a competition \textcolor{brown}{\textless task type\textgreater}. The previous solution contains a bug. According to the buggy information, revise it to fix the issue.
            
            \textbf{\textcolor{blue}{Task description: }}
            \textcolor{brown}{\textless task description\textgreater}

            \textbf{\textcolor{blue}{Evaluation metric: }}
            \textcolor{brown}{\textless metric\textgreater}

            \textbf{\textcolor{blue}{Training set format: }}
            \textcolor{brown}{\textless dataset description\textgreater}

            \textbf{\textcolor{orange}{Submission format: }}
            \textcolor{brown}{\textless submission format\textgreater}

            \textbf{\textcolor{green}{History Information: }}
            \textcolor{brown}{\textless history information\textgreater}

            \textbf{\textcolor{green}{Previous Solution: }}
            \textcolor{brown}{\textless previous solution\textgreater}

            \textbf{\textcolor{purple}{Previous (buggy) Implementation: }}
            \textcolor{brown}{\textless previous (buggy) implementation \textgreater}

            \textbf{\textcolor{purple}{Previous (buggy) Output: }}
            \textcolor{brown}{\textless previous (buggy) output \textgreater}

            \textbf{\textcolor{blue}{Instructions: }}
            \textcolor{brown}{\textless Response Format\textgreater}, \textcolor{brown}{\textless Implementation Guideline\textgreater}, \textcolor{orange}{\textless Solution Draft Sketch Guideline\textgreater}, \textcolor{green}{\textless Solution Improvement Sketch Guideline\textgreater}, \textcolor{purple}{\textless Solution Debug Sketch Guideline\textgreater}
            
        }
        \end{AIbox}
    \end{minipage}
    \begin{minipage}{\textwidth}
        \centering
        \begin{AIbox}{OPT-Agent-NP}
        {   
            \textbf{\textcolor{orange}{Introduction (draft): }}
            You are a great expert solving \textcolor{brown}{\textless task description\textgreater} question. You should propose a solution to this question.
            
            \textbf{\textcolor{green}{Introduction (improve): }}
            You are a great expert solving \textcolor{brown}{\textless task description\textgreater} question. You should optimize the solution based on the history information.
            
            \textbf{\textcolor{purple}{Introduction (debug): }}
            You are a great expert solving \textcolor{brown}{\textless task description\textgreater} question. You should debug the solution based on the previous buggy information.
            
            \textbf{\textcolor{blue}{Task description: }}
            \textcolor{brown}{\textless task description\textgreater}

            \textbf{\textcolor{blue}{Submission Format: }}
            \textcolor{brown}{\textless submission format\textgreater}

            \textbf{\textcolor{blue}{Question: }}
            The \textcolor{brown}{\textless task type\textgreater} question is: \textcolor{brown}{\textless question\textgreater}

            \textbf{\textcolor{green}{History Information: }}
            \textcolor{green}{\textless history information\textgreater}

            \textbf{\textcolor{purple}{Previous Buggy Information: }}
            \textcolor{purple}{\textless Previous buggy information\textgreater}

            \textbf{\textcolor{blue}{Example Input and Output: }}
            \textcolor{brown}{\textless Example Input and Output\textgreater}

            \textbf{\textcolor{blue}{Instructions: }}
            \textcolor{brown}{\textless Instructions\textgreater}

            \textbf{\textcolor{blue}{Response Format: }}
            \textcolor{brown}{\textless Response Format\textgreater}
            
        }
        \end{AIbox}
        \caption{\textbf{Prompt Template of OPT-Agent.} \textcolor{orange}{Orange denotes draft action.} \textcolor{green}{Green denotes improve action.} \textcolor{purple}{Purple denotes debug action.} \textcolor{blue}{Blue denotes shared prompts.}}

        \label{fig:agent_ml_np}
    \end{minipage}
\end{figure*}

\subsection{OPT-Agent Results Analysis}
\begin{figure*}[htbp]
    \centering
    \includegraphics[width=\linewidth]{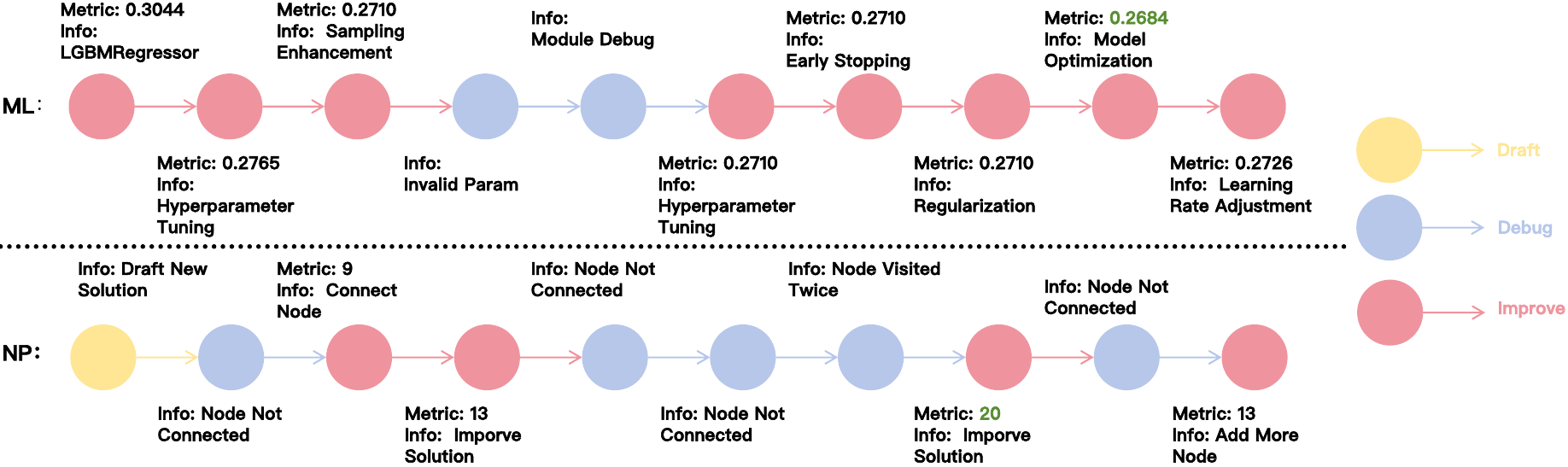}
    \caption{
    \textbf{Optimization Trajectories.} 
    The figure contrasts the agent's path against the environment. In ML (Top), the agent uses error logs to monotonicially improve, demonstrating true \textit{self-optimization}. In NP (Bottom), feedback often triggers erratic jumps, indicating a struggle to map discrete environmental signals to valid solution updates.
    }
    \label{fig:agent-trace}
    \vspace{-1em}
\end{figure*}
\subsection{ML Task}
\begin{figure*}[ht]
    \centering
    \includegraphics[width=0.9\linewidth]{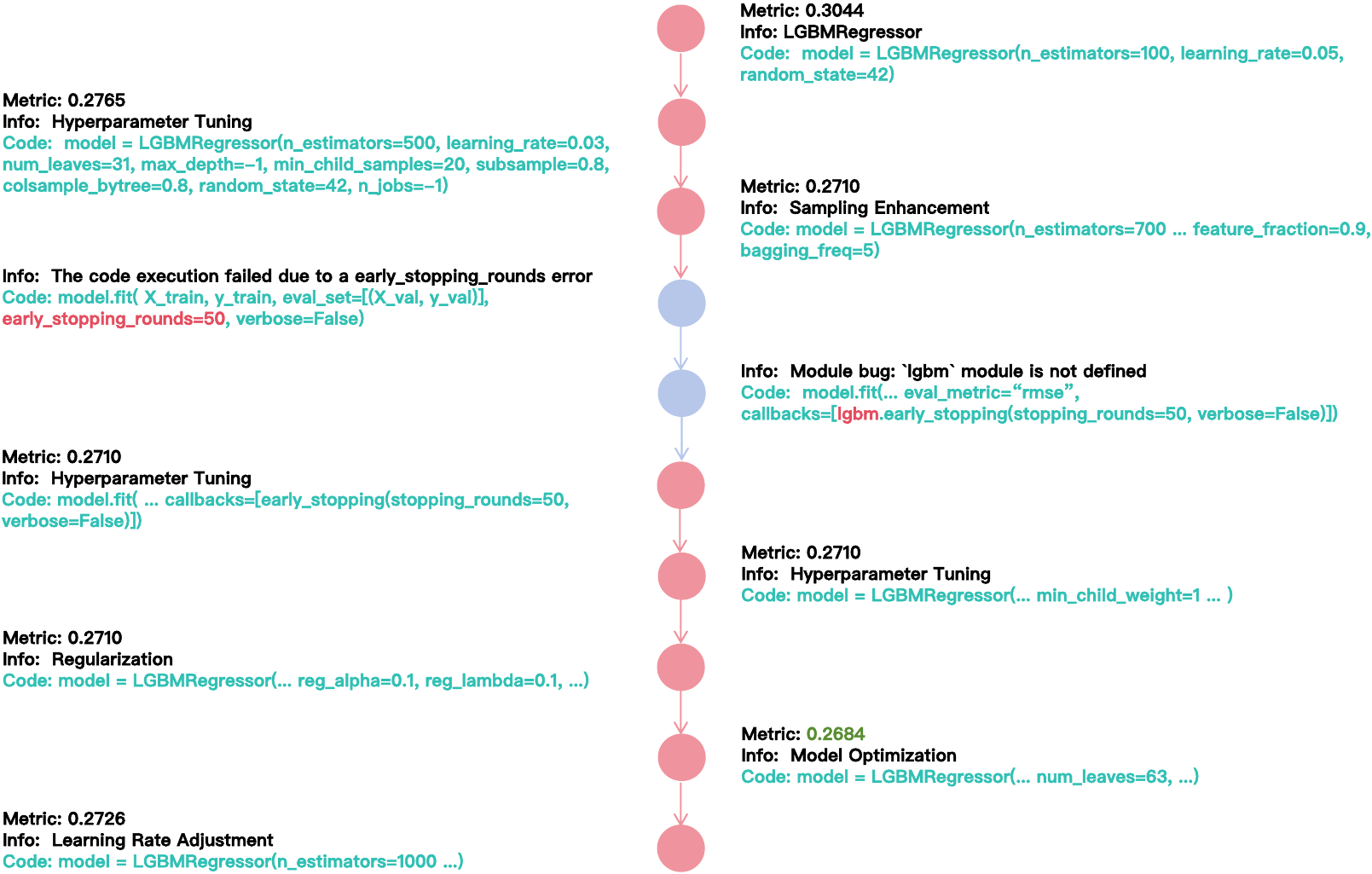}
    \caption{\textbf{Detailed OPT-Agent-ML Trace on the Bike Sharing Demand Task}, utilizing \texttt{gemini-2.0-flash} as LLM base model. The red, and blue nodes represent the improve, and debug action, respectively.}
    \label{fig:app_ml_case}
\end{figure*}

As shown in Figure~\ref{fig:app_ml_case}, we present the optimization trajectory of OPT-Agent tackling bike sharing demand ML task, illustrating progressive improvements in evaluation metrics through various strategies. Beginning with hyperparameter tuning and sampling enhancements, the model undergoes iterative refinements including the introduction of early stopping, regularization techniques, and learning rate adjustments. The diagram also highlights encountered issues, such as the early stopping rounds exception, along with the corresponding fixes, demonstrating a systematic approach to model optimization and performance enhancement.
\subsection{NP Problem}
\begin{figure*}[ht]
    \centering
    \includegraphics[width=0.9\linewidth]{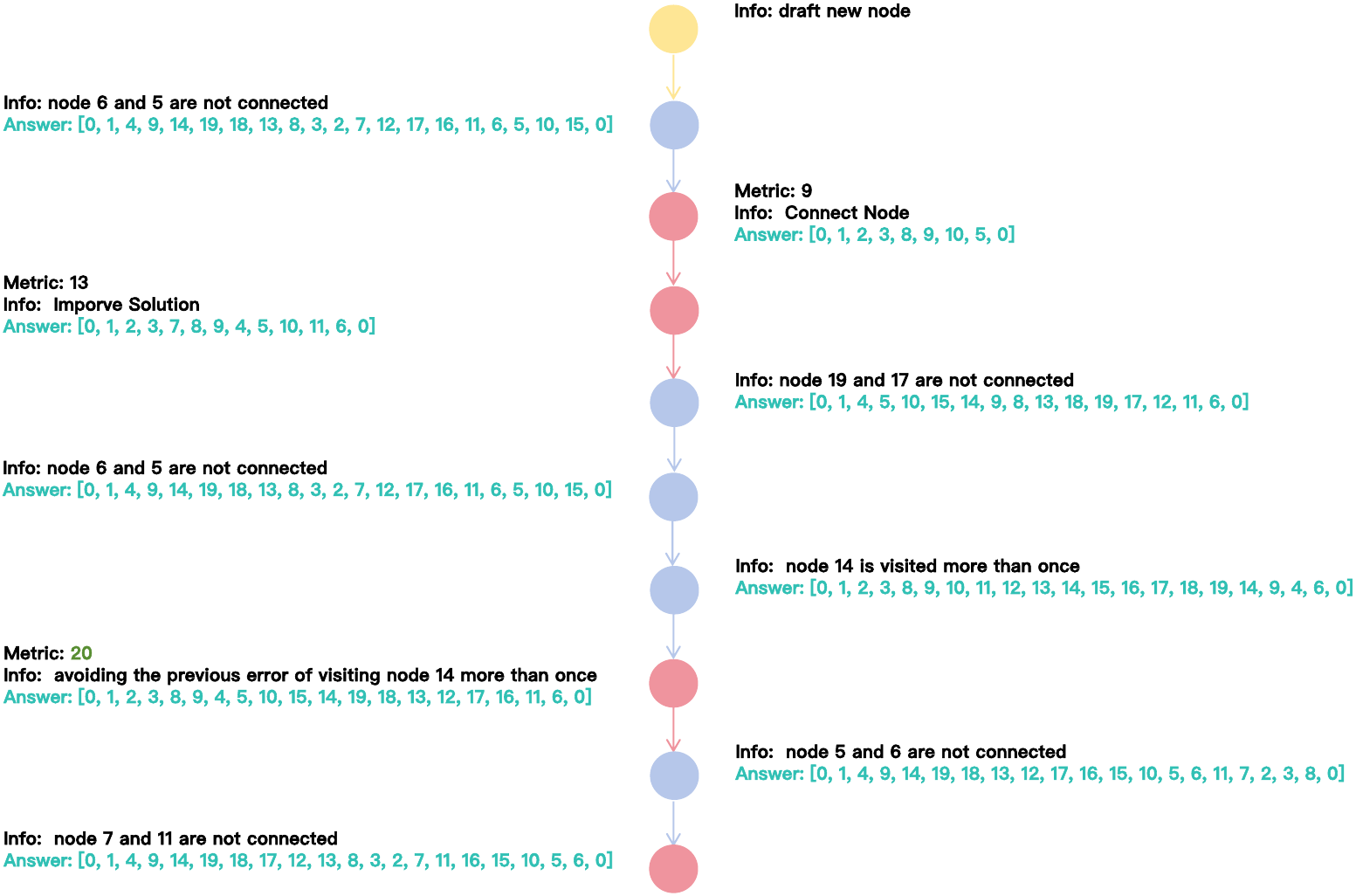}
    \caption{\textbf{Detailed OPT-Agent-NP Trace on the Hamiltonian Cycle Task}, utilizing \texttt{gemini-2.0-flash} as the LLM base model. The yellow, red, and blue nodes represent the draft, improve, and debug action, respectively.}
    \label{fig:app_np_case}

\end{figure*}

As shown in Figure~\ref{fig:app_np_case}, we illustrate the solution refinement process of OPT-Agent applied to the Hamiltonian Cycle NP problem. The flowchart depicts iterative attempts to build a valid Hamiltonian circuit by resolving challenges such as disconnected nodes and repeated visits. Each step includes metric evaluations, detailed state information, and proposed paths, demonstrating how OPT-Agent systematically enhances the solution toward a valid and optimized Hamiltonian cycle.

\begin{figure*}[ht]
    \centering
    \vspace{-2pt}
    \begin{minipage}{\textwidth}
        \centering
        \begin{AIbox}{OPT-Agent-ML}
        {
            \textbf{\textcolor{blue}{Response format: }} Your response should be a brief outline/sketch of your proposed solution for model, optimizer, and hyperparameters selection in natural language (3-5 sentences), followed by a single markdown code block (wrapped in ```) which implements this solution and prints out the evaluation metric. There should be no additional headings or text in your response. Just natural language text followed by a newline and then the markdown code block. Note that the code block should be a complete Python program. 

            \textbf{\textcolor{blue}{Impl guideline: }} Be aware of the running time of the code, it should complete within \texttt{time}. All data is already available in the \texttt{./input} directory. You can also use the \texttt{./working} directory to store any temporary files that your code needs to create. The evaluation should be based on \texttt{k-fold-validation} but only if that’s an appropriate evaluation for the task at hand.

            \textbf{\textcolor{blue}{Solution draft sketch guideline: }} The initial solution design should be simple, efficient, and avoid overfitting, with minimal iterations. Take the Memory section into consideration when proposing the design, and do not propose the same modeling solution while keeping the evaluation the same. The solution sketch should be 3-5 sentences and propose a reasonable evaluation metric for this task. Do not suggest performing Exploratory Data Analysis (EDA). The data is already prepared and available in the \texttt{./input directory}, so there is no need to unzip any files. Note that the training dataset should be shuffled before splitting into training and validation sets, and the random seed (state) should be fixed.

            \textbf{\textcolor{blue}{Solution improvement sketch guideline: }} The solution sketch should be a brief natural language description of how the previous solution can be improved. You should be very specific and propose only a single actionable improvement. Do not suggest performing Exploratory Data Analysis (EDA). Ensure that function parameters match the official documentation by checking for accuracy, compatibility, and any deprecated or renamed parameters, referring to the latest examples if needed. Note that only the model, optimizer, hyperparameters, and feature engineering should be modified. This improvement should be atomic so that its effect can be experimentally evaluated. Additionally, take the Memory section into consideration when proposing the improvement. The solution sketch should be 3-5 sentences.

            \textbf{\textcolor{blue}{Solution debug sketch guideline: }} You should write a brief natural language description (3-5 sentences) of how the issue in the previous implementation can be fixed. Do not suggest performing Exploratory Data Analysis (EDA). Ensure that function parameters match the official documentation by checking for accuracy, compatibility, and any deprecated or renamed parameters, referring to the latest examples if needed. If the previous buggy solution was due to time limitations, focus on reducing the code’s time consumption rather than fixing the bug—for example, by simplifying the model’s hyperparameters, reducing the number of iterations, or switching from K-Fold cross-validation to a single train-test split. Additionally, take the Memory section into consideration when proposing the improvement.
        }
        \end{AIbox}
        \vspace{2em}
    \end{minipage}

            
            

    \caption{\textbf{Fixed prompts in OPT-Agent}. This encompasses the response format, implementation guidelines, solution draft sketch guidelines, solution improvement sketch guidelines, and solution debug sketch guidelines for ML tasks, as well as example inputs and outputs, instructions, and response format for NP problems.}
    \label{fig:prompt_ml_np}
\end{figure*}

\end{document}